\pdfoutput=1
\documentclass[11pt]{article}

\usepackage{authblk}
\usepackage[]{acl}

\usepackage{times}
\usepackage{latexsym}

\usepackage[T1]{fontenc}
\usepackage[utf8]{inputenc}

\usepackage{microtype}
\usepackage{booktabs}
\usepackage{xcolor}
\usepackage{graphicx}
\usepackage{pgfplots}
\usepackage{subfig}
\makeatletter \renewcommand\AB@affilsepx{\qquad \protect\Affilfont} \makeatother
\title{Self-Supervised Losses for One-Class Textual Anomaly Detection}
\author[1,2]{\textbf{Kimberly T. Mai}}
\author[1,2]{\textbf{Toby Davies}}
\author[1]{\textbf{Lewis D. Griffin}}
\affil[1]{University College London}\makeatletter \renewcommand\AB@affilsepx{\protect\Affilfont} \makeatother
\affil[2]{The Alan Turing Institute\authorcr}
\affil[ ]{\texttt{\{kimberly.mai, toby.davies, l.griffin\}@ucl.ac.uk}}

\begin{document}
\maketitle
\begin{abstract}
Current deep learning methods for anomaly detection in text rely on supervisory signals in inliers that may be unobtainable or bespoke architectures that are difficult to tune. We study a simpler alternative: fine-tuning Transformers on the inlier data with self-supervised objectives and using the losses as an anomaly score. Overall, the self-supervision approach outperforms other methods under various anomaly detection scenarios, improving the AUROC score on semantic anomalies by 11.6\% and on syntactic anomalies by 22.8\% on average. Additionally, the optimal objective and resultant learnt representation depend on the type of downstream anomaly. The separability of anomalies and inliers signals that a representation is more effective for detecting semantic anomalies, whilst the presence of narrow feature directions signals a representation that is effective for detecting syntactic anomalies.
\end{abstract}

\section{Introduction}
Anomaly detection is the task of identifying unusual samples relative to an exemplar inlier distribution. It has numerous applications in natural language processing (NLP), including fake news detection \citep{lee-etal-2021-towards}, spam detection \citep{crawford2015survey}, and flagging atypical reviews \citep{ruff2019}.

The difficulty of anomaly detection depends on the magnitude of difference between an anomalous representation and the distribution of inlier representations. Existing works in NLP focus on the far out-of-distribution (OOD) setting \cite{winkens2020contrastive} in which the anomalies are derived from a distinct dataset \citep{hendrycks2020pretrained, arora2021types, Li2021kFoldenKE, podolskiy2021revisiting, zhou2021contrastive}. For example, a model is trained on a sentiment classification dataset, and then that model is used to identify news articles as anomalies. These approaches also often assume the model is trained to classify the distinct inlier sub-classes. The anomaly scoring mechanisms typically leverage these supervisory signals by fitting a Mahalanobis distance \citep{lee2018simple} to each sub-class or by obtaining the highest probability in the softmax layer \citep{hendrycks17baseline}. However, these supervisory signals may not always be available.

As an alternative configuration, we analyse the one-class anomaly detection setting on more challenging near-OOD anomalies. One-class anomaly detection assumes only inlier data are available at training time and only have one label. Instead of supervisory signals, we study the performance of fine-tuning a Transformer on the inlier data using various self-supervised objectives, and we use the loss as the anomaly score. We examine anomaly detection performance on two near-OOD anomaly types: semantic anomalies, which are created by partitioning a single dataset by class label, and syntactic anomalies, which are created by randomly shuffling inlier sentences. We find that fine-tuning on a pre-trained Transformer outperforms existing and more complex methods, boosting AUROC score on semantic anomalies by 11.6\% and on syntactic anomalies by 22.8\% on average.  

Our findings also suggest that the separation of anomalies and inlier classes in the learnt representation space of the detectors is a strong signal for detecting semantic anomalies, whilst adversarially brittle features are a better indicator of performance in the syntactic anomaly detection setting. Overall, our results indicate the fine-tuning paradigm is a simple baseline that can achieve good results, and the self-supervised objectives used for fine-tuning exploit different cues to identify anomalies. 
\section{Approach}

\subsection{Models}
Using the loss of a fine-tuned Transformer for anomaly detection is analogous to using an autoencoder's reconstruction error as an anomaly score in vision \citep{sakurada2014anomaly}. We anticipate that the fine-tuned models can learn the underlying characteristics of inlier data but not those of anomalies. Hence, the loss is used as the anomaly score as it should be higher for anomalous instances.

We analyse three self-supervised objectives in our experiments. To minimise the influence of architectural differences, we use the encoder from a pre-trained uncased BERT\textsubscript{BASE} \citep{devlin-etal-2019-bert} and append different heads depending on the objective. We fine-tune each model for a maximum of 30,000 steps on inlier data, employing early stopping based on the inlier validation set's loss.

\textbf{Masked language modelling (MLM).} We retain the default configuration for BERT\textsubscript{BASE} and randomly mask 15\% of tokens. At inference time, we mask the same proportion of tokens in the test sentences and use the error between the predicted and true tokens as the anomaly score.

\textbf{Causal language modelling (CLM).} We fine-tune the model to predict the next token given previous tokens in the sequence and use perplexity as the anomaly score. Perplexity has been used to evaluate evidence-supported fact-checking \citep{lee-etal-2021-towards} and far-OOD detection \cite{arora2021types}. Our work differs as it uses perplexity to evaluate more difficult anomalies and does not require auxiliary data.

\textbf{Contrastive loss (SimCSE).} Previous works in vision suggest a contrastive loss can help discriminate anomalies from inliers \citep{tack2020csi, sehwag2021ssd}. However, these methods require data augmentations that are not directly transferrable to NLP.

SimCSE \cite{gao2021simcse} resolves the data augmentation issue by applying different dropout masks to sentences and trains the model to select the same sentence from a minibatch of other sentence pairs. We fine-tune the model using the default dropout probability ($p=0.1$) and temperature ($\tau=0.05$) described in SimCSE and evaluate anomalies using the NT-Xent loss \cite{pmlr-v119-chen20j}.

We compare the three fine-tuned models to four baselines:

\textbf{Pre-trained BERT (Pre-trained).} We evaluate MLM loss on BERT\textsubscript{BASE}  without any fine-tuning. This configuration can be compared to MLM to examine the incremental benefit of fine-tuning. We disregard the auxiliary next-sentence prediction objective as we do not use sentence pairs for anomaly detection.

\textbf{Other attention-based anomaly detectors.} We compare our approach to two state-of-the-art methods which use attention.
CVDD \citep{ruff2019} learns a set of compact context vectors to describe the inlier data using a multi-head self-attention mechanism. It evaluates a sentence through the average cosine distance of the sentence's contextual embedding to the context vectors.

DATE \cite{date-manolache-2021} adapts ELECTRA \citep{clark2020} for the anomaly detection task. DATE includes an additional objective to predict which pre-defined pattern was used by the generator to mask the input tokens. At inference time, the input text is fed into the discriminator directly. The average probability of each token being uncorrupted serves as the anomaly score.

\textbf{Bag-of-words models (BoW).} We follow the approach in CVDD and compute the mean over word embeddings extracted from FastText \cite{bojanowski-etal-2017-enriching} to create a sentence embedding for each datum. We use these sentence embeddings to train linear OC-SVMs, which worked better than using $k$-NNs or Mahalanobis distances in our experiments.

\begin{figure*}[t]%
    \centering
    \subfloat[\centering Semantic anomaly results.]{{\includegraphics[width=0.5\linewidth]{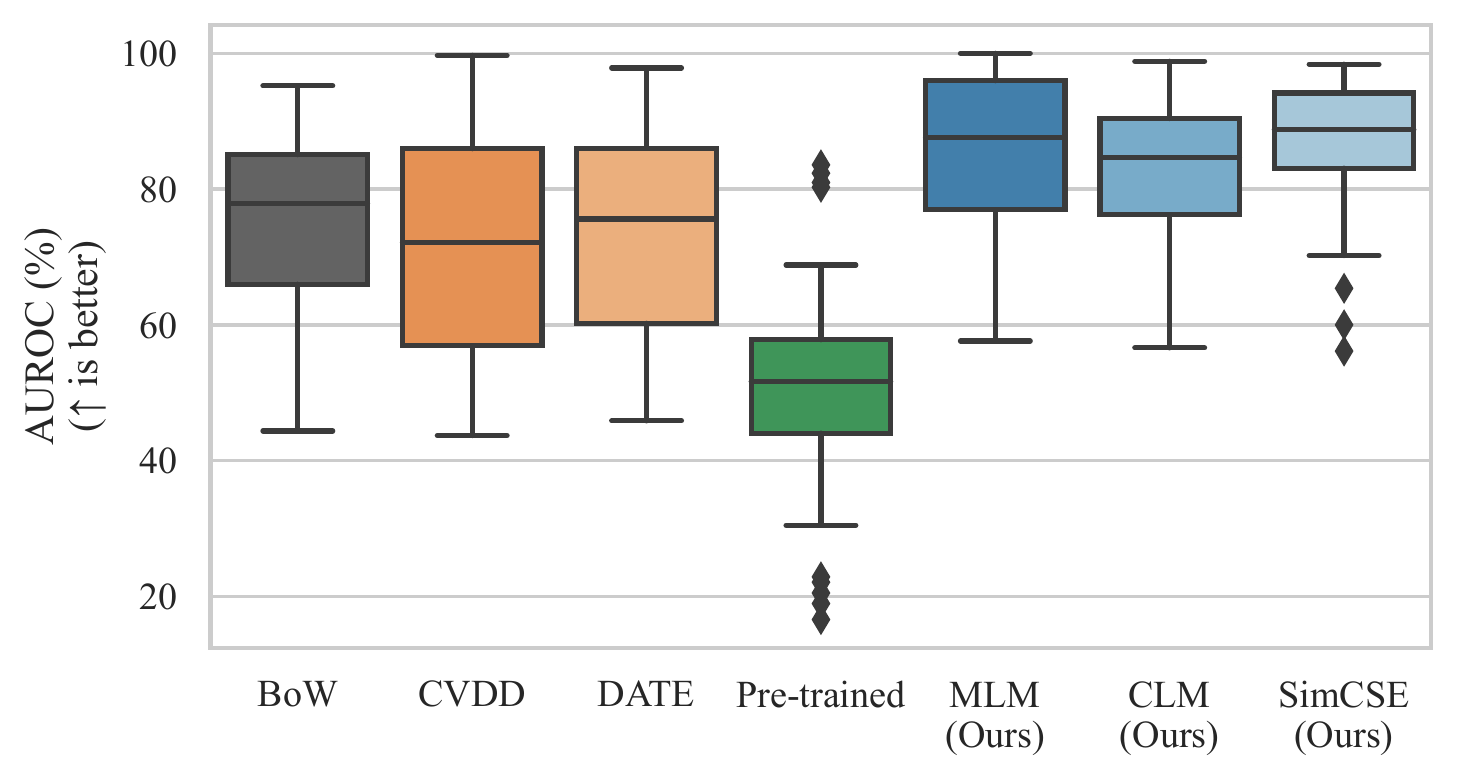} }}%
    \subfloat[\centering Syntactic anomaly results encompassing all $n$-grams.]{{\includegraphics[width=0.5\linewidth]{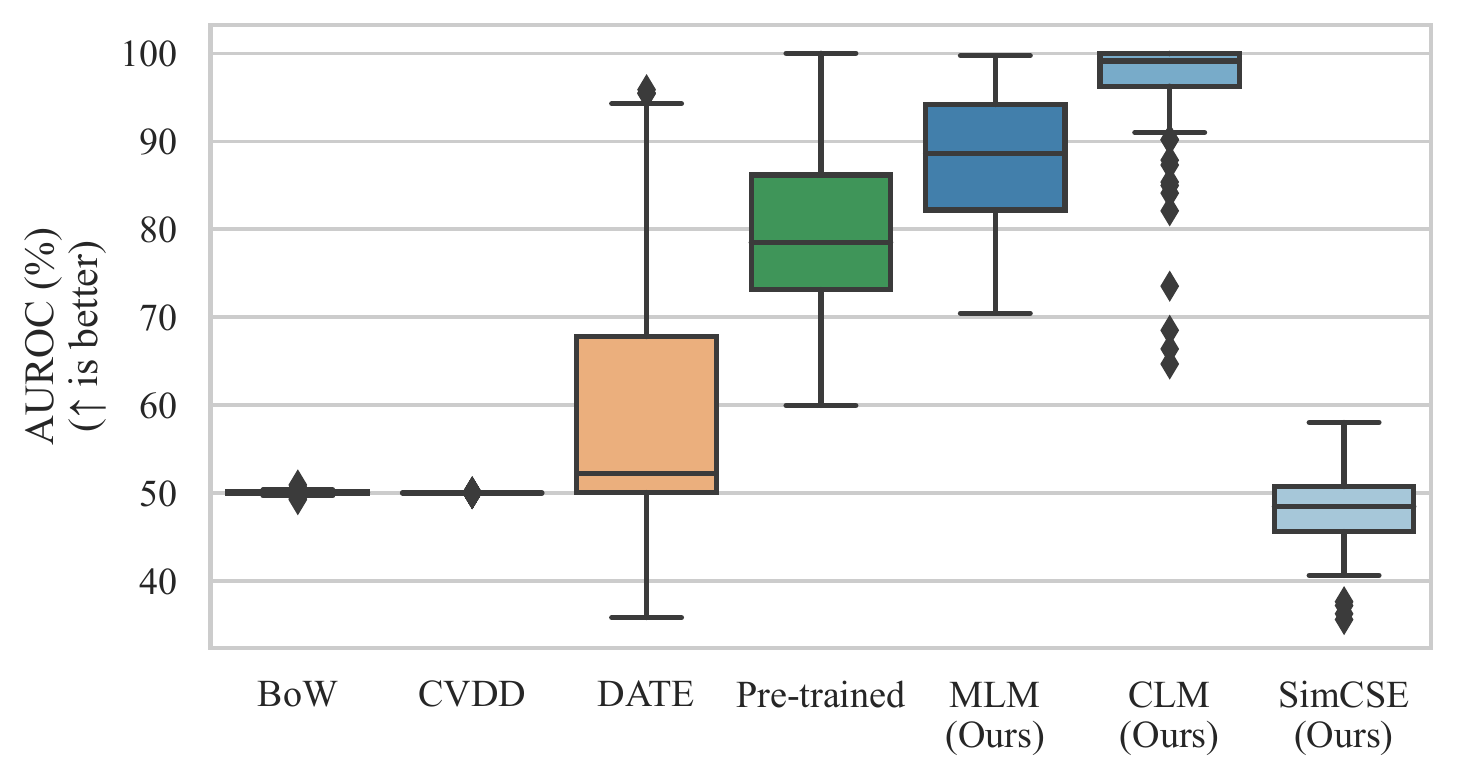} }}%
    \caption{Anomaly detection results aggregated by model.}%
    \label{fig:overall res}%
\end{figure*}
\subsection{Datasets and anomaly detection setup}
\label{sec:datasets}
To allow comparison with the baseline methods, we evaluate anomaly detection performance on 20 Newsgroups \citep{lang95}, Reuters-21578 \citep{Lewis1997Reuters21578TC}, AG News \citep{Zhang2015CharacterlevelCN} and IMDb Movie Reviews \citep{maas-EtAl:2011:ACL-HLT2011}. We also perform experiments on Snopes \cite{vo2020facts} (a fact-checking dataset) and the Enron Spam Dataset \cite{metsis2006} to simulate more realistic anomaly detection applications. We pre-process each dataset by lowercasing text, stripping punctuation and removing stopwords as per \citet{ruff2019}.

We use the datasets' class labels to construct two setups for the inlier training data. This allows us to examine anomaly detection performance in the settings where the inliers are narrow and more diverse. For a dataset with $m$ class labels:
\begin{itemize}
    \item Unimodal normality: We construct the inliers using data from a single label.
    \item Multimodal normality: We construct the inliers using data from $m-1$ labels.
\end{itemize}
\begin{table}[h]
\centering
\small
\begin{tabular}{ll} 
\toprule
\textbf{Class}      & \textbf{Sentence}                          \\ 
\midrule
Inlier & \scriptsize{\colorbox{teal!10}{voip} \colorbox{teal!20}{gaining} \colorbox{teal!30}{ground} \colorbox{teal!40}{despite} \colorbox{teal!50}{cost} \colorbox{teal!60}{concerns}}   \\
Anomaly &  \scriptsize{\colorbox{teal!60}{concerns} \colorbox{teal!10}{voip} \colorbox{teal!40}{despite} \colorbox{teal!50}{cost} \colorbox{teal!30}{ground} \colorbox{teal!20}{gaining}} \\
\bottomrule
\end{tabular}
\caption{Example of a syntactic anomaly derived from the AG News dataset. We look at $n$-grams ($n \in \{1,2,3,4\}$) and shuffle them until each $n$-gram is no longer in its original position.}
\label{fig:permutation}
\end{table}

We use the test splits of each dataset to formulate two types of near-OOD anomalies:
\begin{itemize}
    \item Semantic anomalies: Data belonging to the same original class label(s) as the training data are categorised as inliers whilst the remainder are categorised as anomalies. %move the description to the figure or remove entirely
    \item Syntactic anomalies: Inlier and anomaly data are derived from the same class of data used to construct the training set. Inlier data are unchanged; anomalies have shuffled word order. To create the anomalies, we implement the seeded random function algorithm in \citet{sinha-etal-2021-masked}. This setup allows us to measure the anomaly detectors' sensitivity to the underlying syntactic information whilst fixing the word frequency statistics. We illustrate an example of a syntactic anomaly in Table \ref{fig:permutation}. 
\end{itemize}

\section{Results}

Figure \ref{fig:overall res} shows the overall anomaly detection results for both types of anomalies. The full results split by dataset and normality are in Appendix \ref{sec:appendix}.

\textbf{Fine-tuning a pre-trained Transformer boosts anomaly detection performance.} In the case of semantic anomalies, although the BoW performance suggests anomaly detection can be performed through analysing word frequency statistics, fine-tuning helps to give additional information about the nature of inliers. This observation aligns with observations in vision \citep{fort2021exploring}. Our approach also outperforms CVDD and DATE, particularly in the multimodal normality setting. 

Fine-tuning also improves syntactic anomaly detection, where frequency statistics are insufficient for discrimination. SimCSE is an exception, and we attribute this to the NT-Xent loss considering the entire sentence representation at inference.

\begin{figure}[h]
    \centering
    \includegraphics[width=\linewidth]{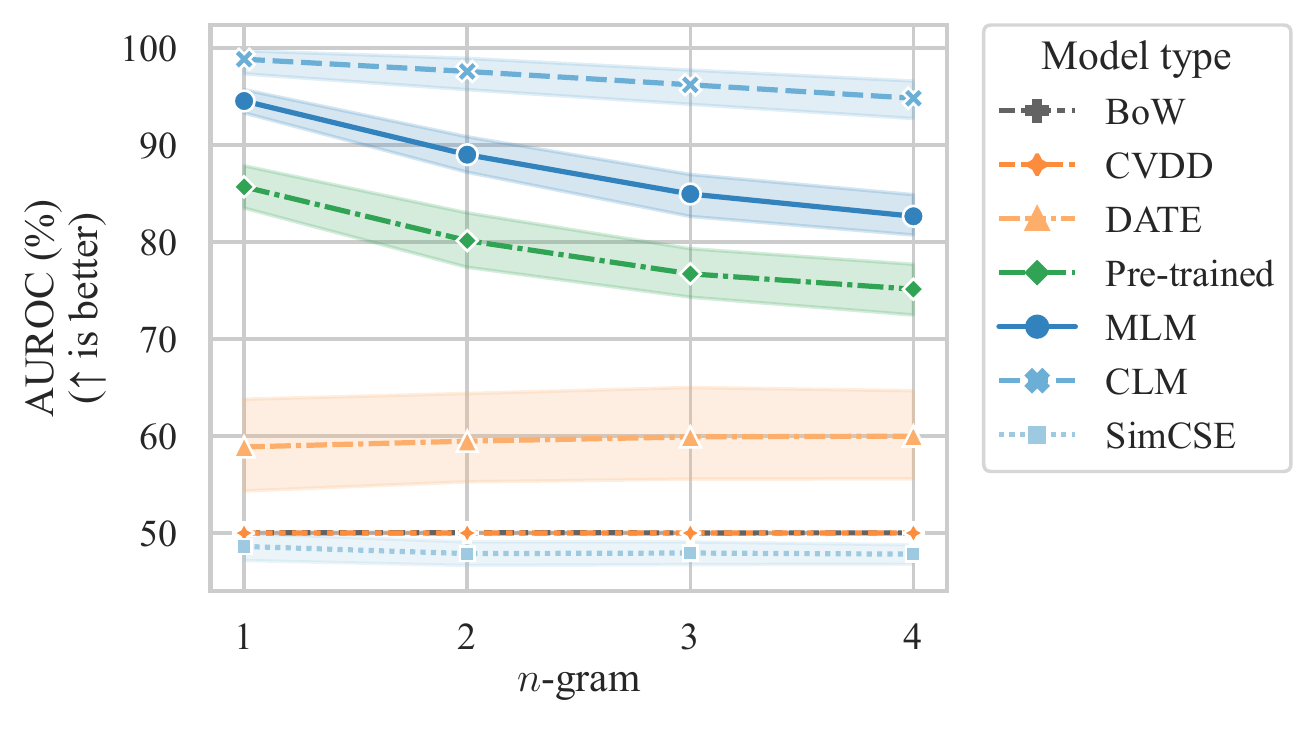}
    \caption{Mean AUROC across datasets on syntactic anomalies by $n$-gram level. Larger $n$-grams are more challenging to differentiate from inliers as fewer individual tokens are shuffled.}
    \label{fig:permutation_ablation}
\end{figure}
\textbf{Density models are much better at detecting syntactic anomalies.} We conducted an ablation study of performance under different permutation strengths. CLM is more stable under more challenging anomaly detection conditions (Figure \ref{fig:permutation_ablation}), experiencing a decline of only 4\% between 1-grams and 4-grams. Pre-trained and fine-tuned MLM experience similar drops (11\%), which indicates the choice of objective for anomaly scoring is a core component for performance. As CLM calculates its score at the token level, it is more sensitive to syntactic changes compared to MLM, which considers spans of tokens through its masking mechanism.

In the following experiments, we extracted the embeddings at the last hidden BERT layer and mean-pooled over the positions to analyse the characteristics of the learnt embeddings.

\begin{figure}[h]
    \centering
    \includegraphics[width=\linewidth]{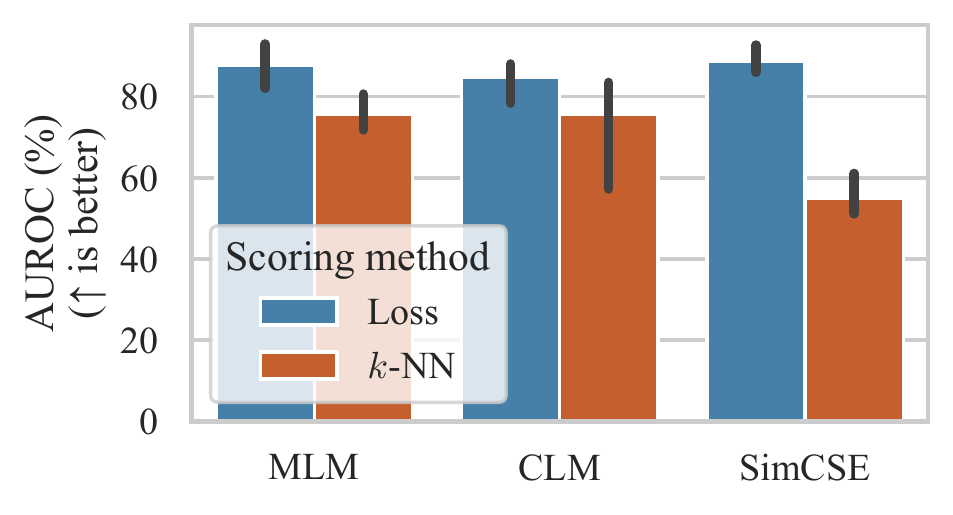}
    \caption{Comparison between using the loss as an anomaly score and $k$-NNs for semantic anomaly detection.}
    \label{fig:raw_embeds}
\end{figure}
\textbf{Using the loss combined with the embedding is better than using the embeddings as a feature extractor.} Figure \ref{fig:raw_embeds} shows the median semantic anomaly detection AUROC score when using the models end-to-end compared to extracting the embedding to train a $k$-NN. Although the raw embeddings are generally capable of performing anomaly detection, end-to-end use of the methods is more discriminative. 

\begin{figure}[h]
    \centering
\includegraphics[width=\linewidth]{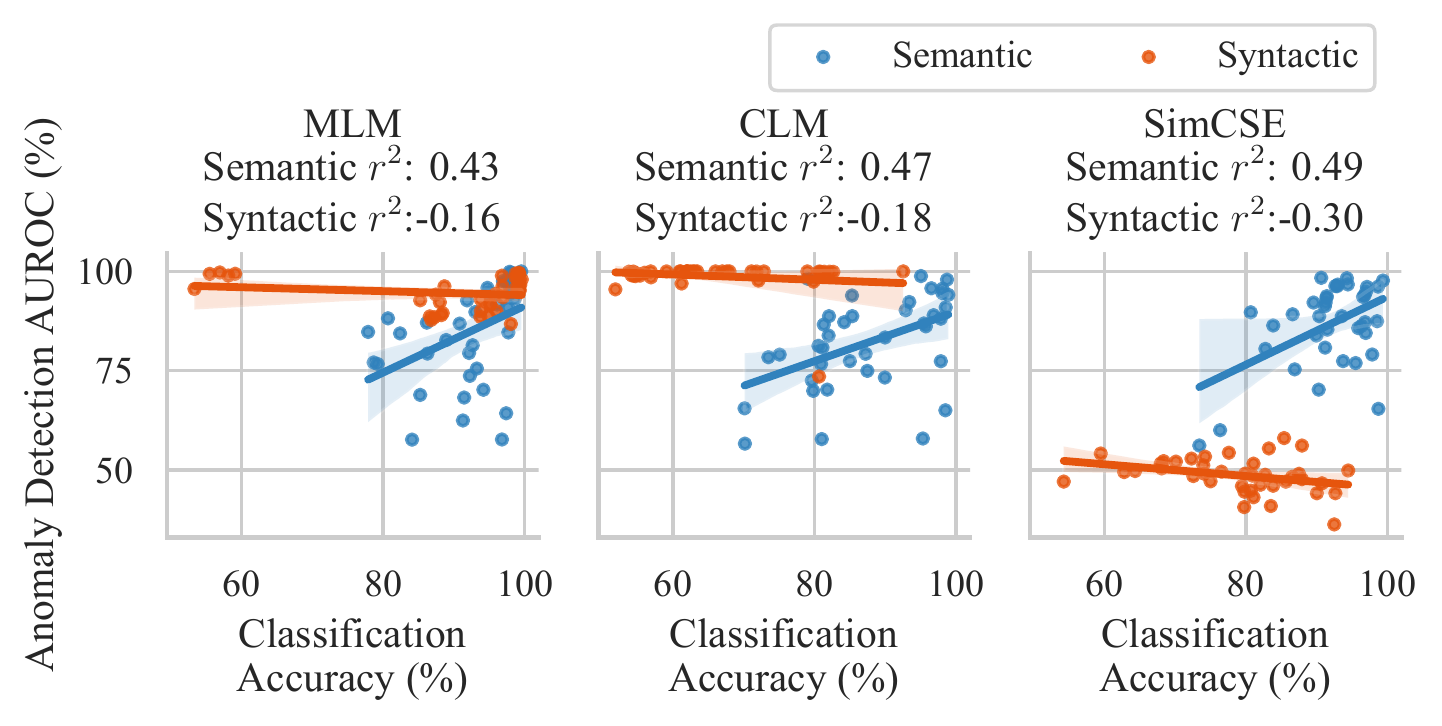}
    \caption{Scatter plot comparing classification accuracy of test inliers versus anomalies to anomaly detection performance across datasets.}
    \label{fig:class_scatter}
\end{figure}
\textbf{Separability of inliers and anomalies is a stronger signal for better semantic anomaly detection.} To examine the separability of embeddings for each learnt representation, we extracted both inlier and anomalous embeddings at the last hidden state and trained a logistic classifier. The correlation between classification accuracy and anomaly detection is more apparent for semantic anomalies (Figure \ref{fig:class_scatter}), suggesting separability is a good indicator for better representations in this case, whereas there is no such relationship for syntactic anomalies. This pattern suggests there is another factor that influences syntactic anomaly detection.

\begin{figure}[h]
    \centering
    \includegraphics[width=\linewidth]{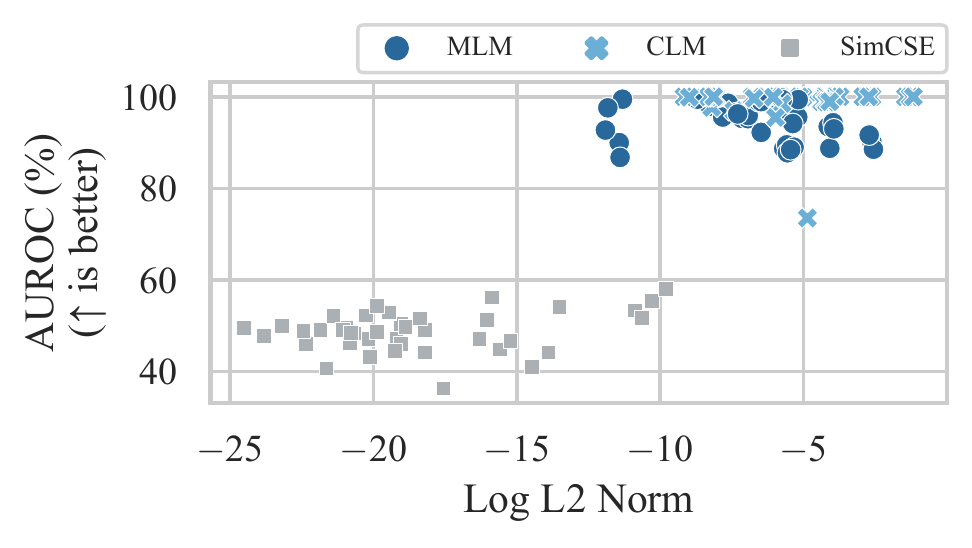}
    \caption{Scatter plot comparing average log L2 norms of the training inlier data to 1-gram syntactic anomaly detection performance. Higher norms are more brittle. The pattern is similar across different $n$-gram levels.}
    \label{fig:norms}
\end{figure}
\textbf{Syntactic anomaly detection performance is more correlated to brittle features.} We hypothesise that a narrower\footnote{Narrow and brittle features refer to non-robust features as defined in adversarial machine learning literature \cite{ilyas2019}.} inlier representation is a better signal for syntactic anomaly performance as it provides more directions for anomalies to manifest.

We adopt the procedure in \citet{mai2021brittle} and calculate the average L2 gradient norms divided by the trace of the covariance matrix with respect to the training data. We observe similar behaviour across all datasets (summarised in Figure \ref{fig:norms}), whereby higher gradient norms clearly correspond to better anomaly detection performance.

Among the methods, CLM-based embeddings tend to be the most brittle and SimCSE the least. This corresponds with previous literature which states that autoregressive models like GPT \cite{radford2018improving} are highly anisotropic \cite{cai2021isotropy}, and models such as SimCSE which are trained on contrastive objectives are more isotropic \cite{wang2020understanding, gao2021simcse}.

\section{Conclusion}
We studied the performance of fine-tuned Transformers using three self-supervised losses through a range of datasets and anomaly detection tasks. We show that this approach outperforms more complex methods, and employing the loss as an anomaly detector is better than using the learnt embeddings as a feature extractor. The best self-supervised loss depends on the nature of the anomalies, which suggests there is scope for analysing ensemble models or outlier exposure in future work.

\section*{Ethical considerations}
Anomaly detectors are practical tools for indicating whether a system is working as intended and for flagging potential hazards \citep{hendrycks2021unsolved}. An adversary may learn how to bypass systems by leveraging anomaly detection research. We restrict this by manually curating inliers and anomalies from publicly available datasets (as described in Section \ref{sec:datasets}). By construction, our experiments are limited to the English language and may not represent features in other languages. We encourage extending our work to other domains and languages to investigate these differences.
\section*{Acknowledgements}
This work was supported by the Engineering and Physical Sciences Research Council (EPSRC) under grant
EP/R513143/1 and The Alan Turing Institute under grant EP/N510129/1. This project made use of time on Tier 2 HPC facility JADE2, funded by
EPSRC under grant EP/T022205/1.

\bibliography{custom}
\bibliographystyle{acl_natbib}

\appendix
\onecolumn
\section{Appendix}
\label{sec:appendix}
\subsection{Semantic anomaly detection results}
%tbc enron
\begin{figure*}[h!]%
    \centering
    \subfloat[\centering AG News (Unimodal)]{{\includegraphics[width=0.35\linewidth]{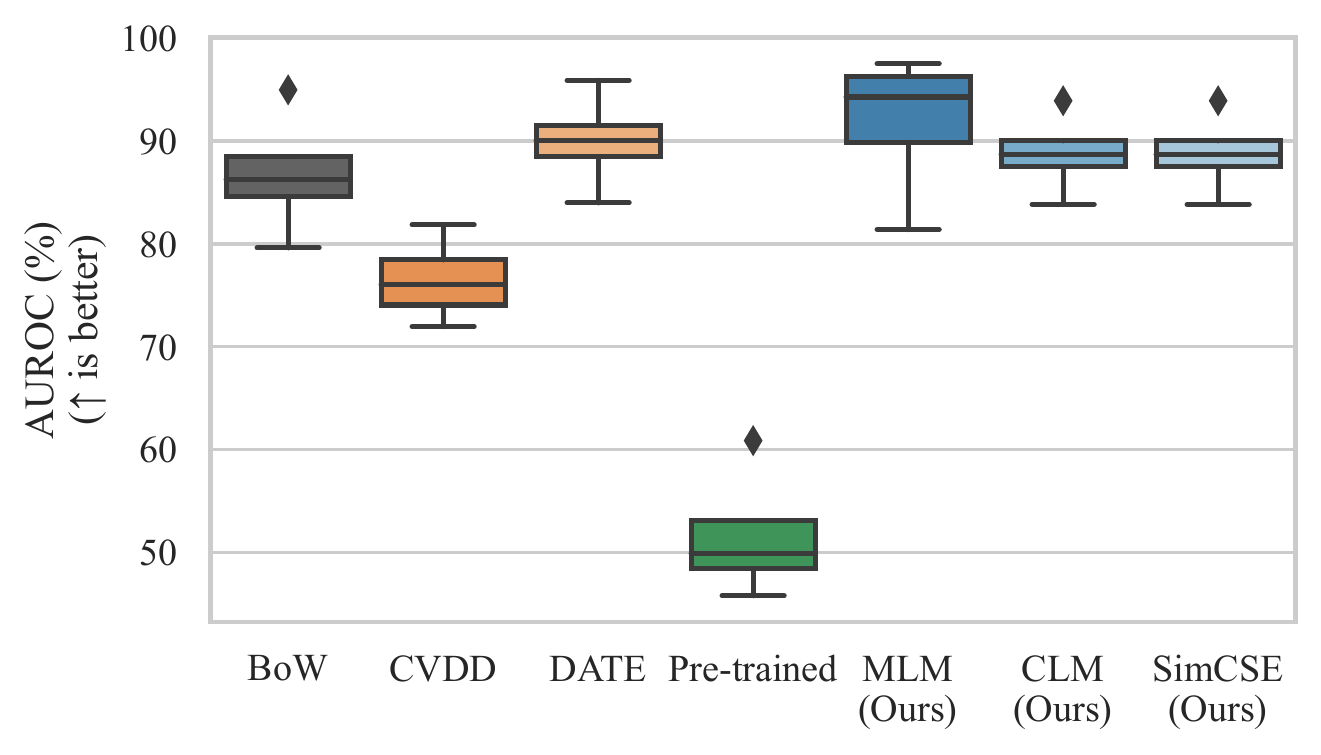} }}% 
    \subfloat[\centering AG News (Multimodal)]{{\includegraphics[width=0.35\linewidth]{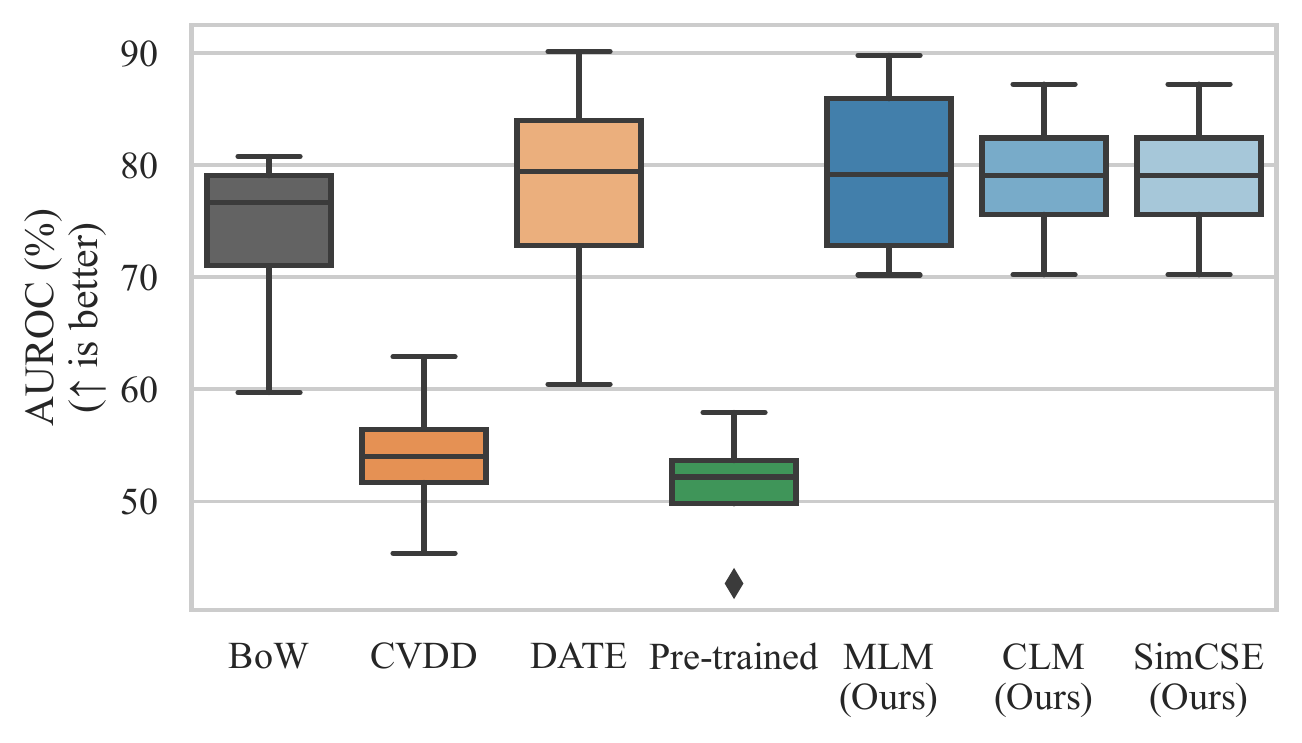} }}\\
    \subfloat[\centering 20 Newsgroups (Unimodal)]{{\includegraphics[width=0.35\linewidth]{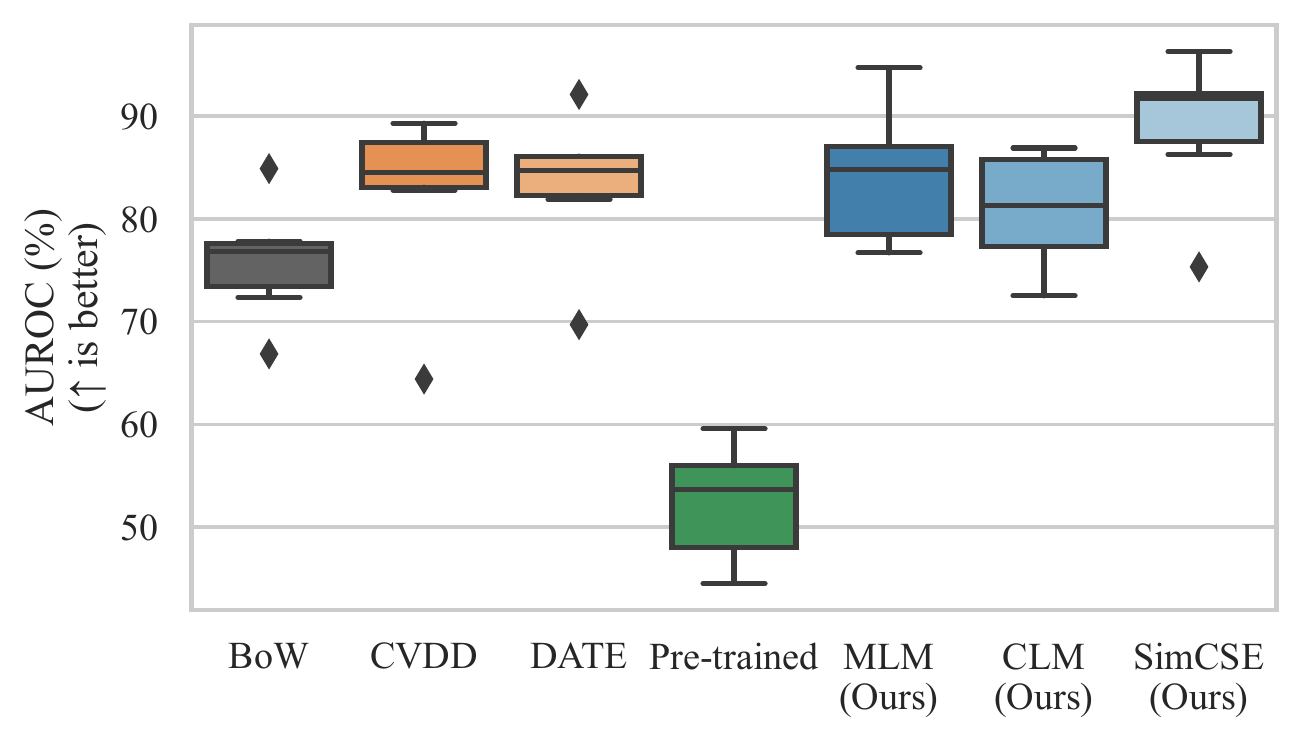} }}
    \subfloat[\centering 20 Newsgroups (Multimodal)]{{\includegraphics[width=0.35\linewidth]{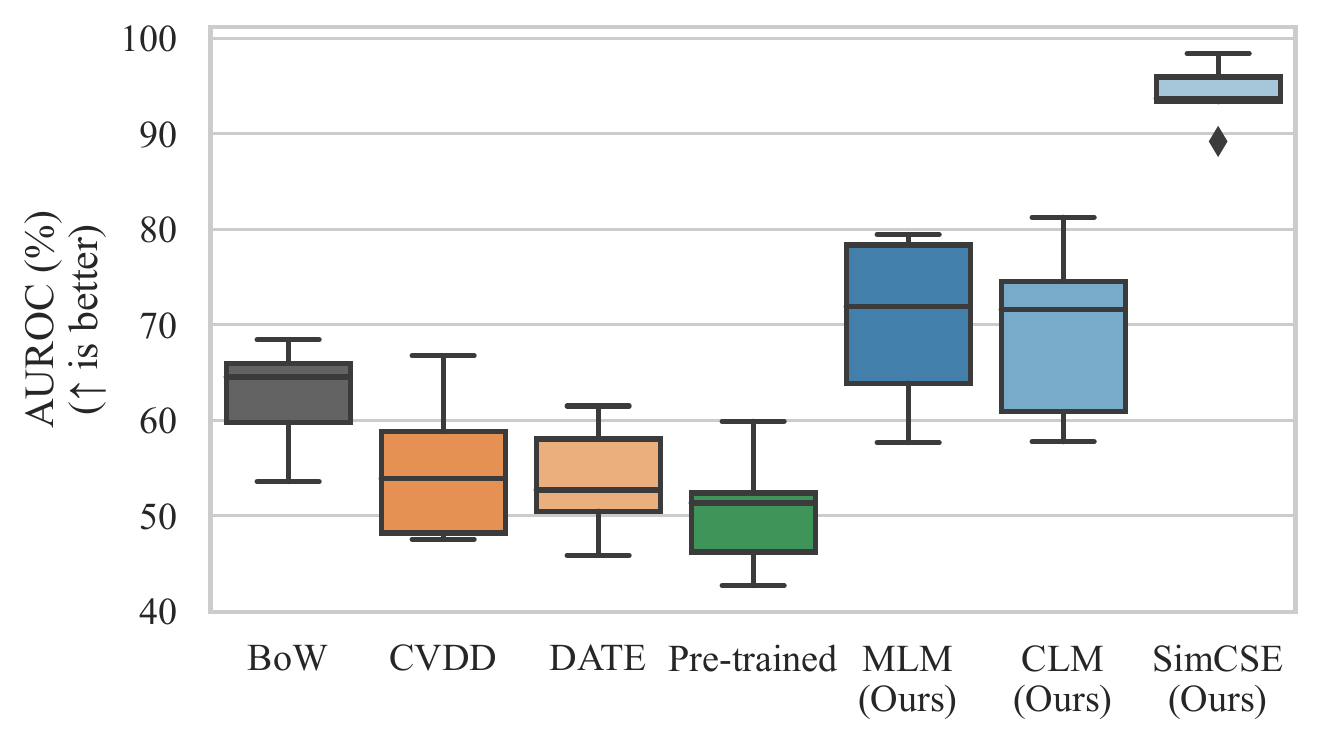} }}\\
    \subfloat[\centering Reuters-21578 (Unimodal)]{{\includegraphics[width=0.35\linewidth]{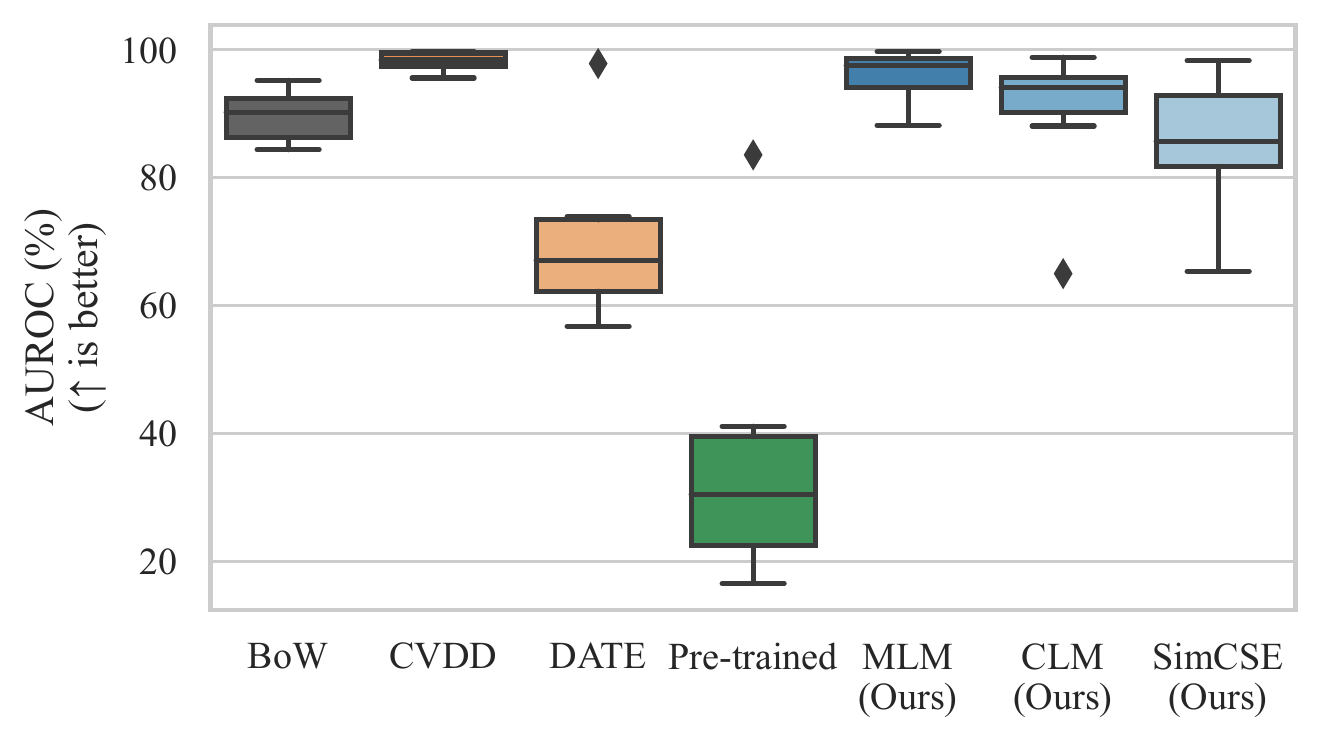} }}
    \subfloat[\centering Reuters-21578 (Multimodal)]{{\includegraphics[width=0.35\linewidth]{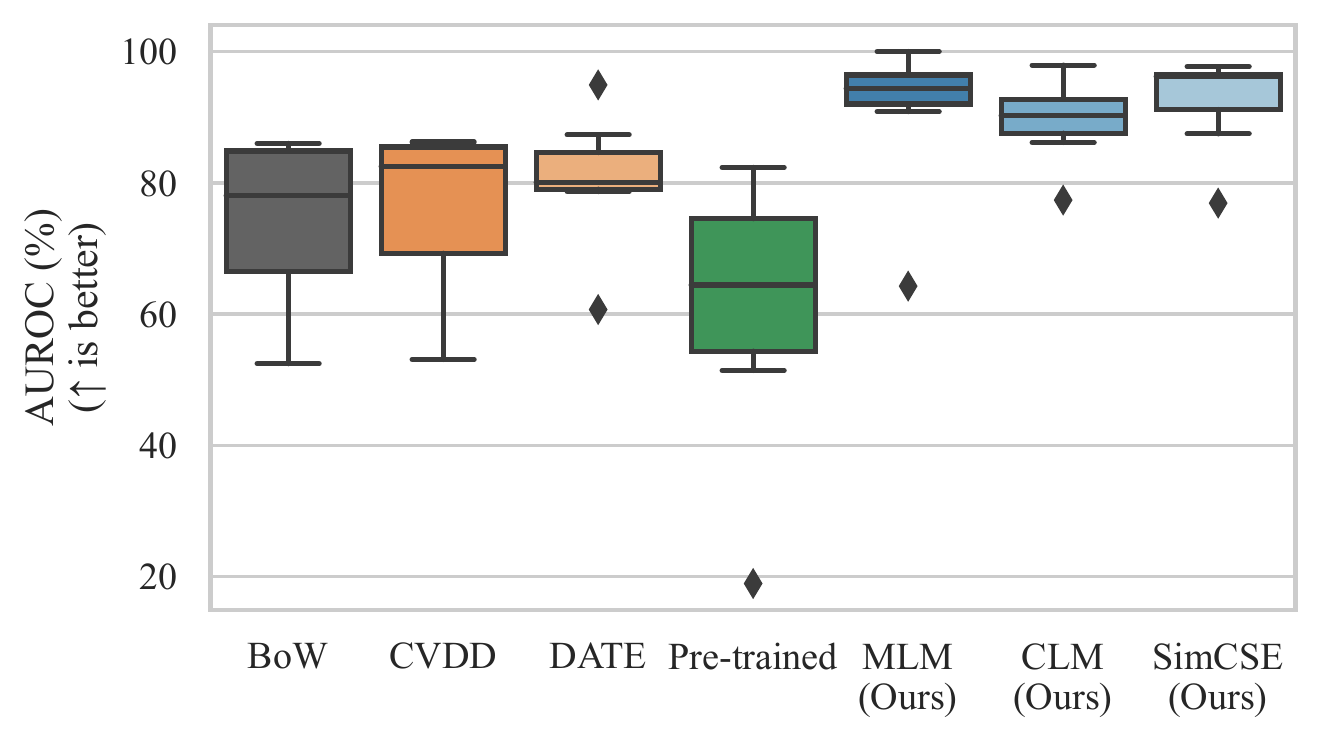} }}\\
    \subfloat[\centering Snopes]{{\includegraphics[width=0.35\linewidth]{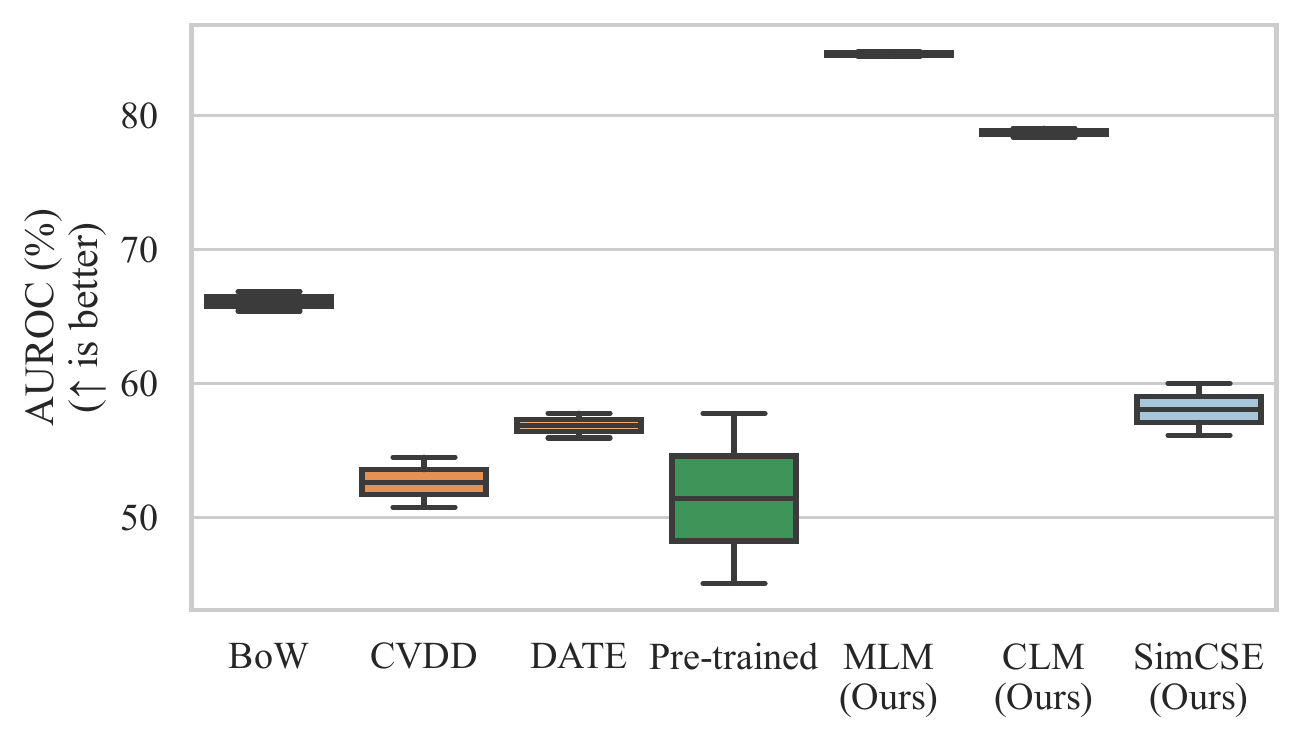} }}
    \subfloat[\centering Enron Spam]{{\includegraphics[width=0.35\linewidth]{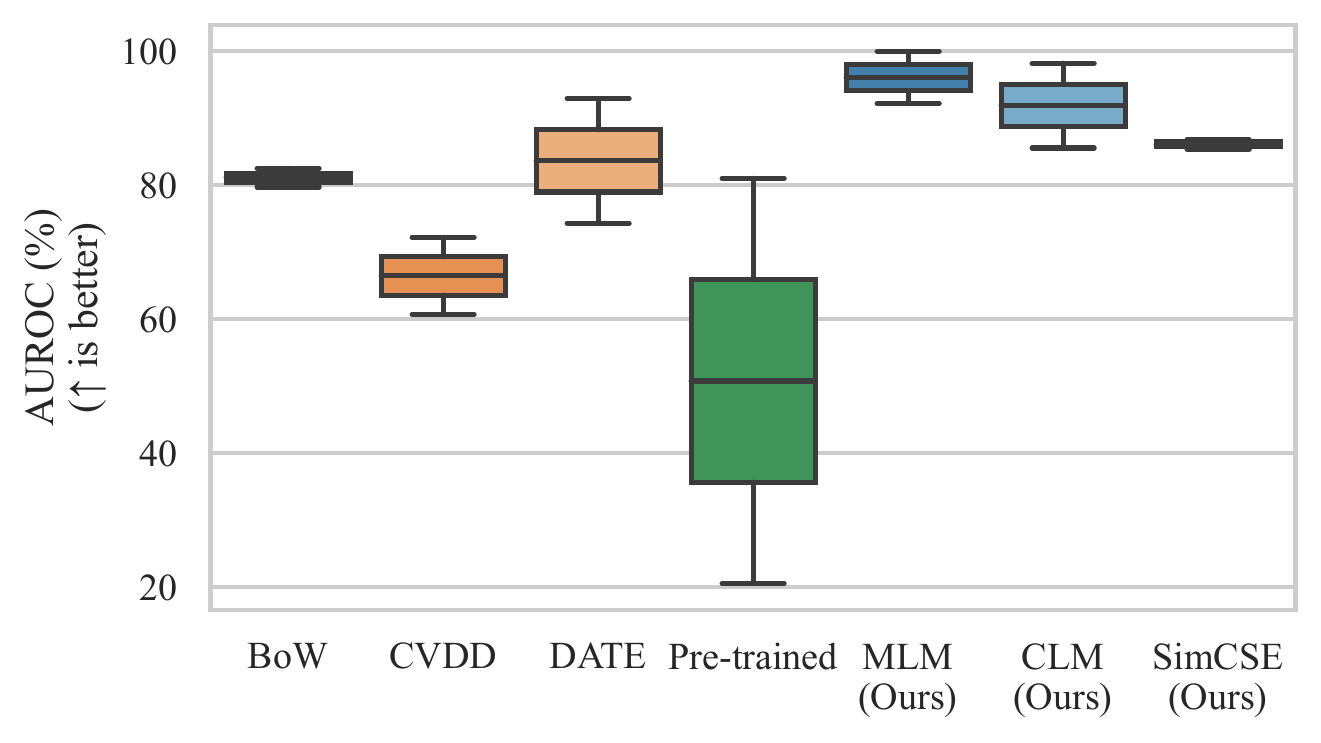} }}\\
    \subfloat[\centering IMDb]{{\includegraphics[width=0.35\linewidth]{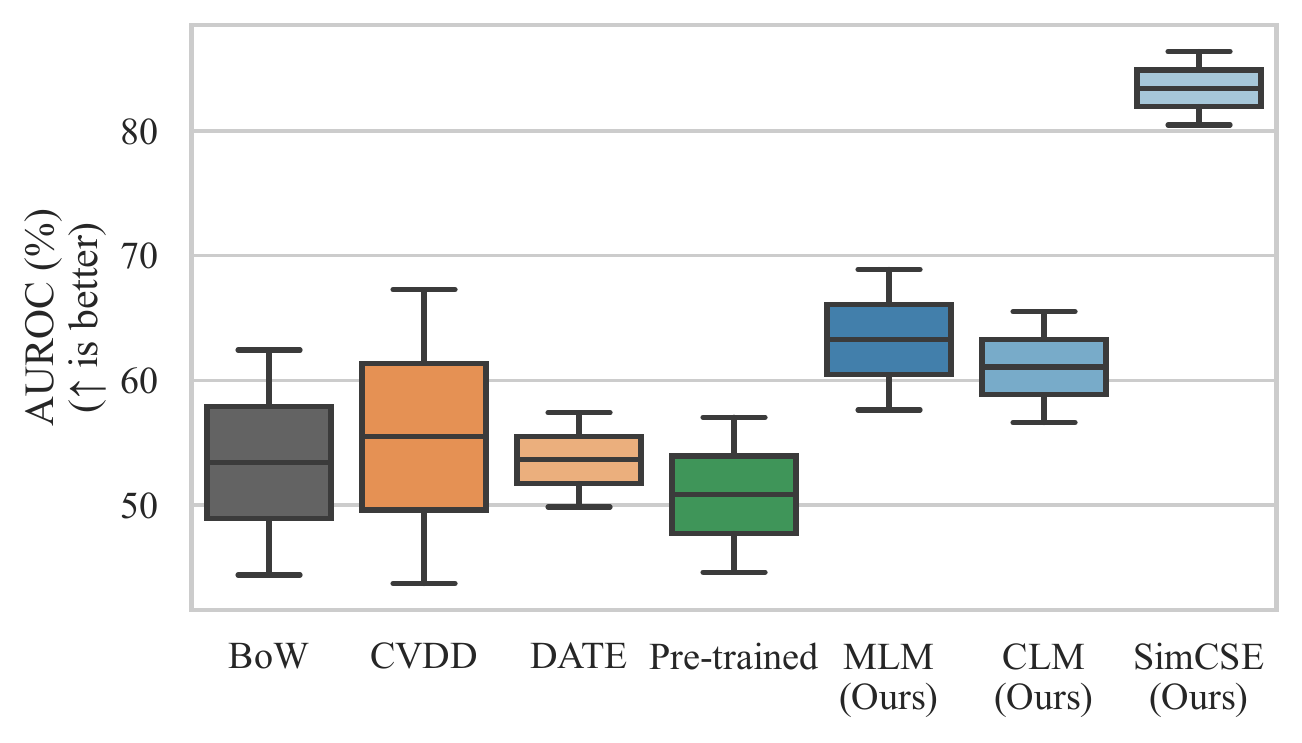} }}\\
    \caption{Semantic anomaly detection results split by dataset.}%
    \label{fig:semantic results}%
\end{figure*}

\newpage
\subsection{Syntactic anomaly detection results}
%tbc enron

\begin{figure*}[h!]%
    \centering
    \subfloat[\centering AG News (Unimodal)]{{\includegraphics[width=0.35\linewidth]{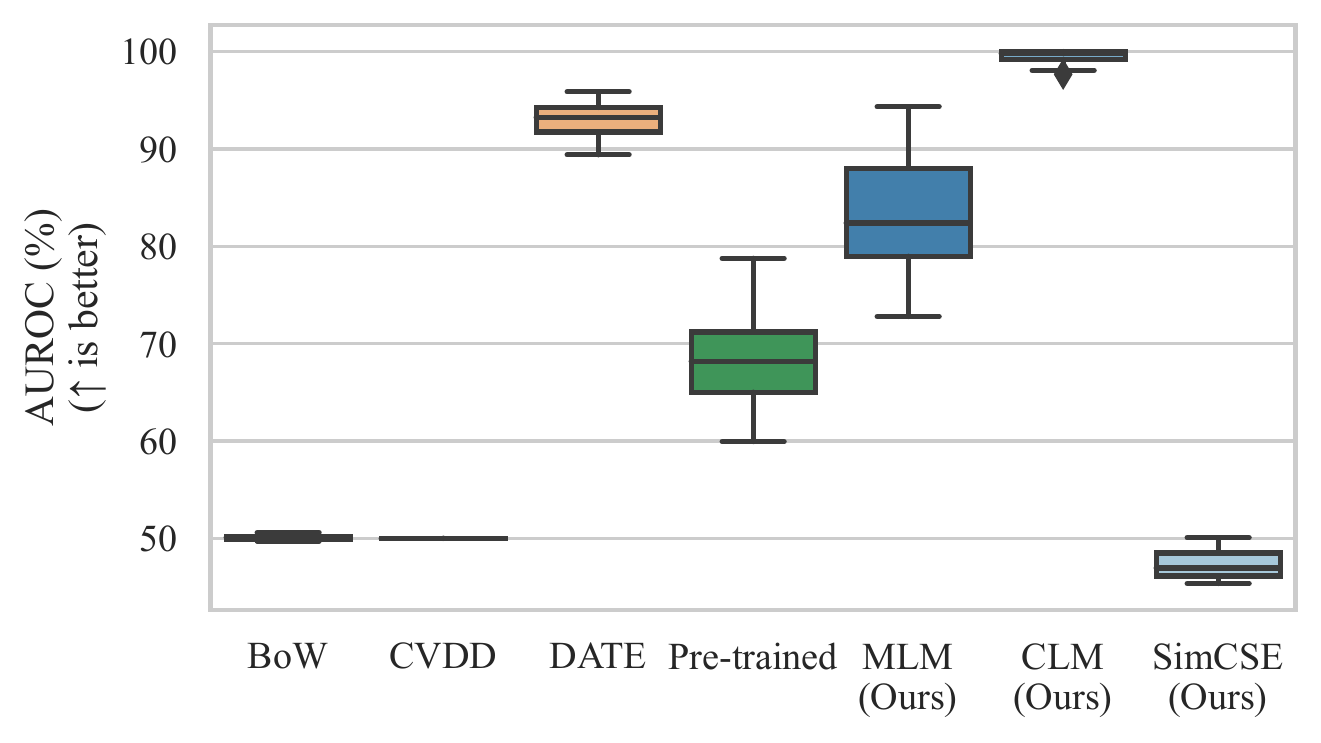} }}% 
    \subfloat[\centering AG News (Multimodal)]{{\includegraphics[width=0.35\linewidth]{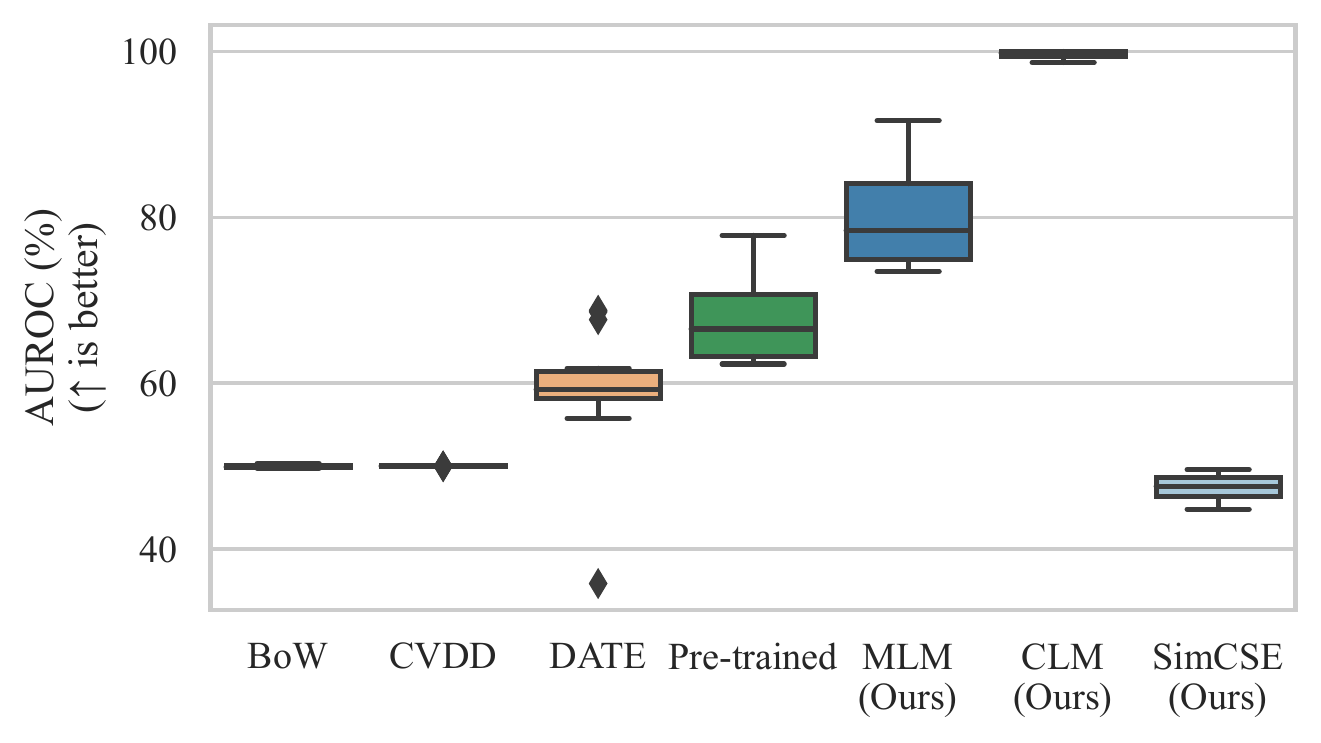} }}\\
    \subfloat[\centering 20 Newsgroups (Unimodal)]{{\includegraphics[width=0.35\linewidth]{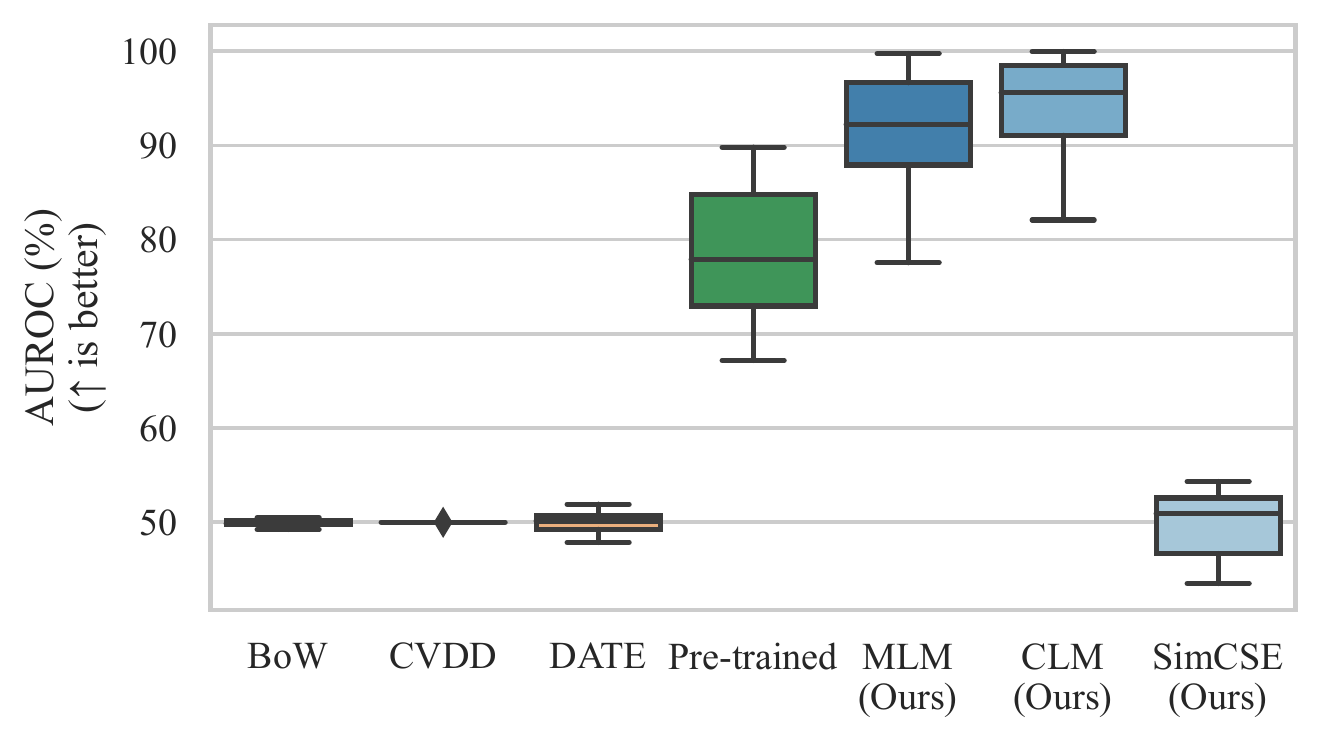} }}
    \subfloat[\centering 20 Newsgroups (Multimodal)]{{\includegraphics[width=0.35\linewidth]{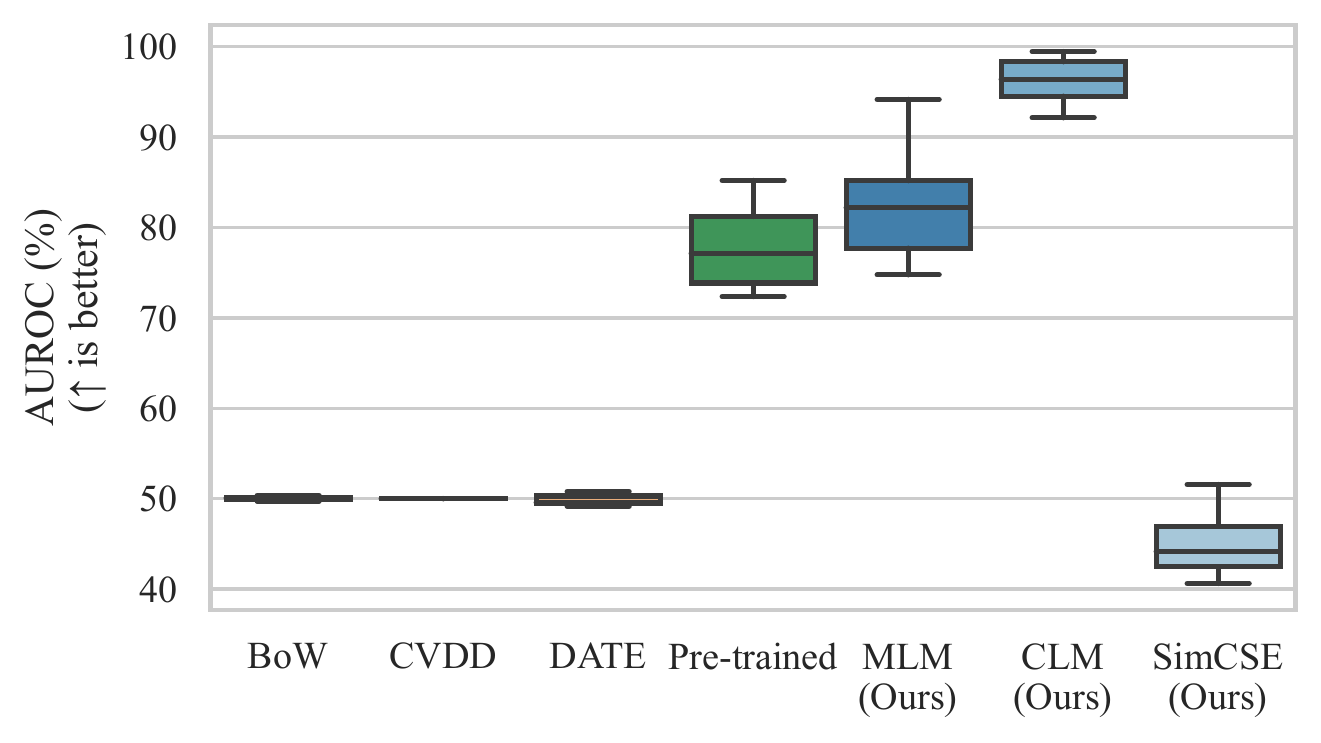} }}\\
    \subfloat[\centering Reuters-21578 (Unimodal)]{{\includegraphics[width=0.35\linewidth]{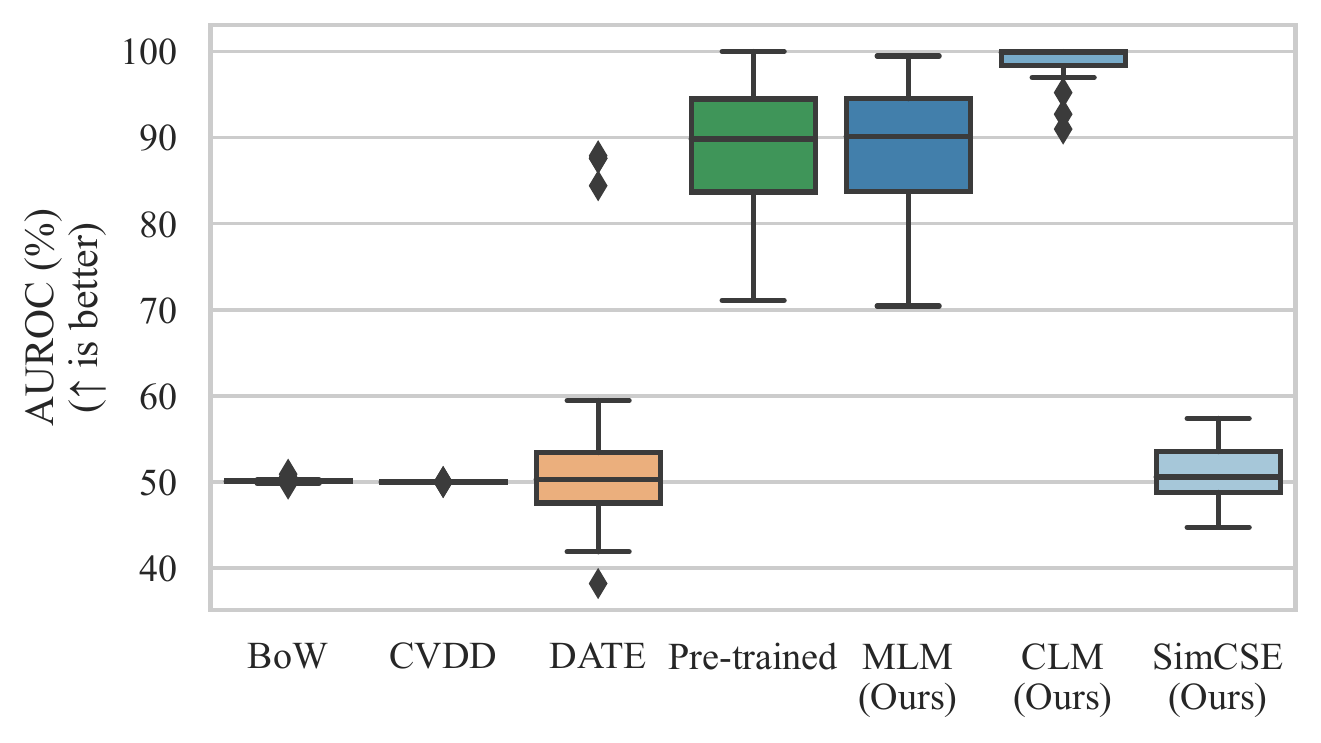} }}
    \subfloat[\centering Reuters-21578 (Multimodal)]{{\includegraphics[width=0.35\linewidth]{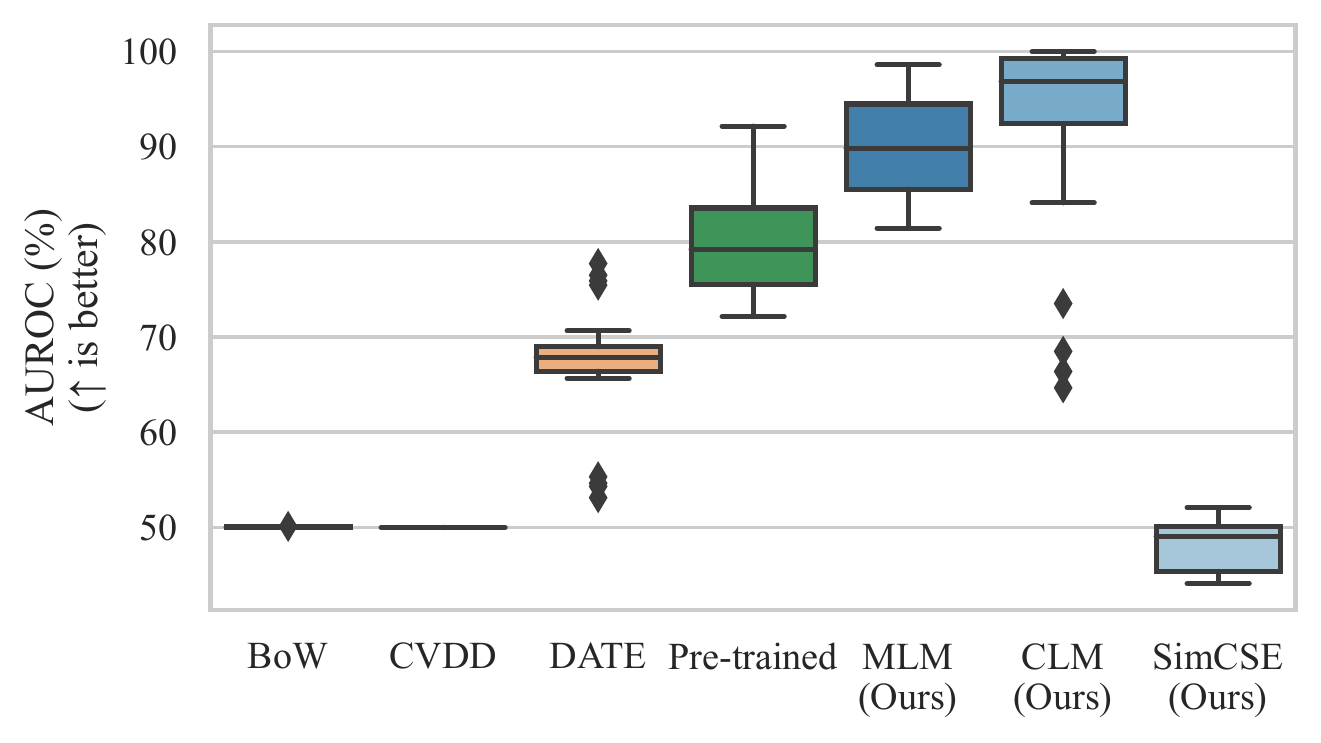} }}\\
    \subfloat[\centering Snopes]{{\includegraphics[width=0.35\linewidth]{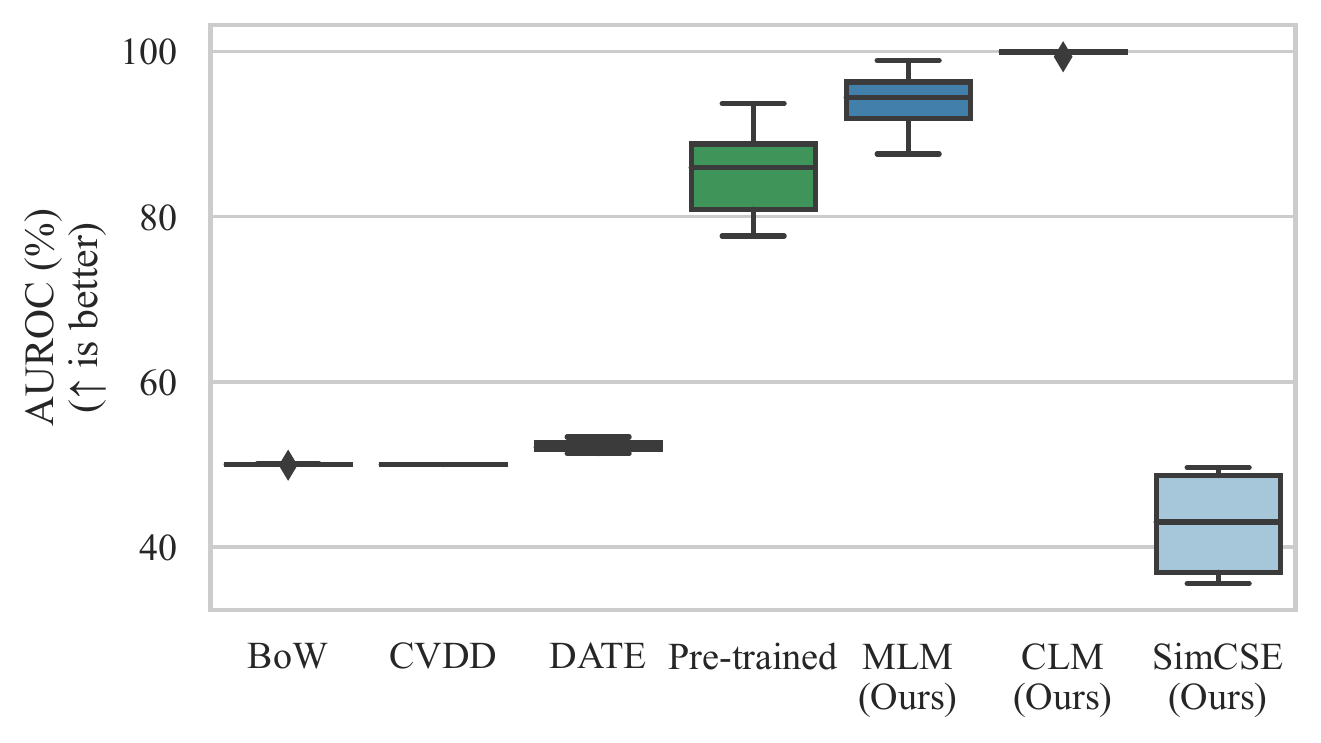} }}
    \subfloat[\centering Enron Spam]{{\includegraphics[width=0.35\linewidth]{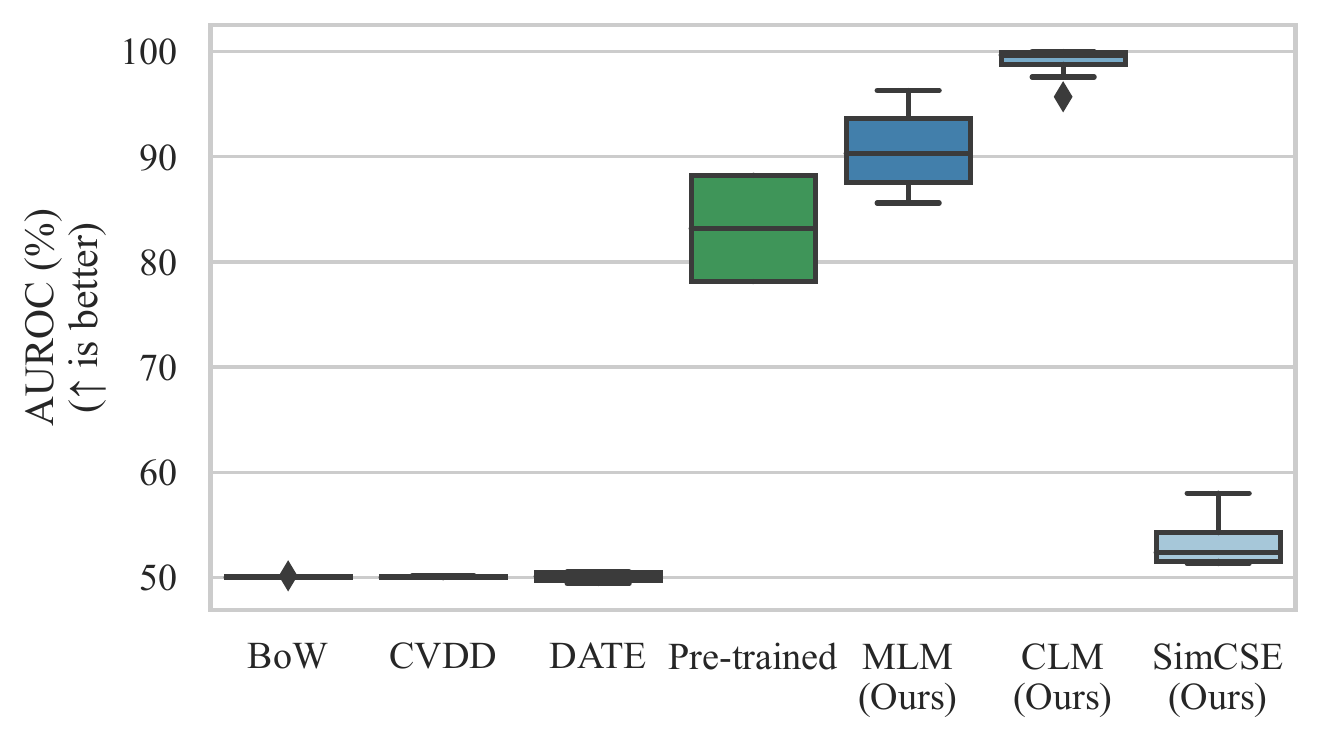} }}\\
    \subfloat[\centering IMDb]{{\includegraphics[width=0.35\linewidth]{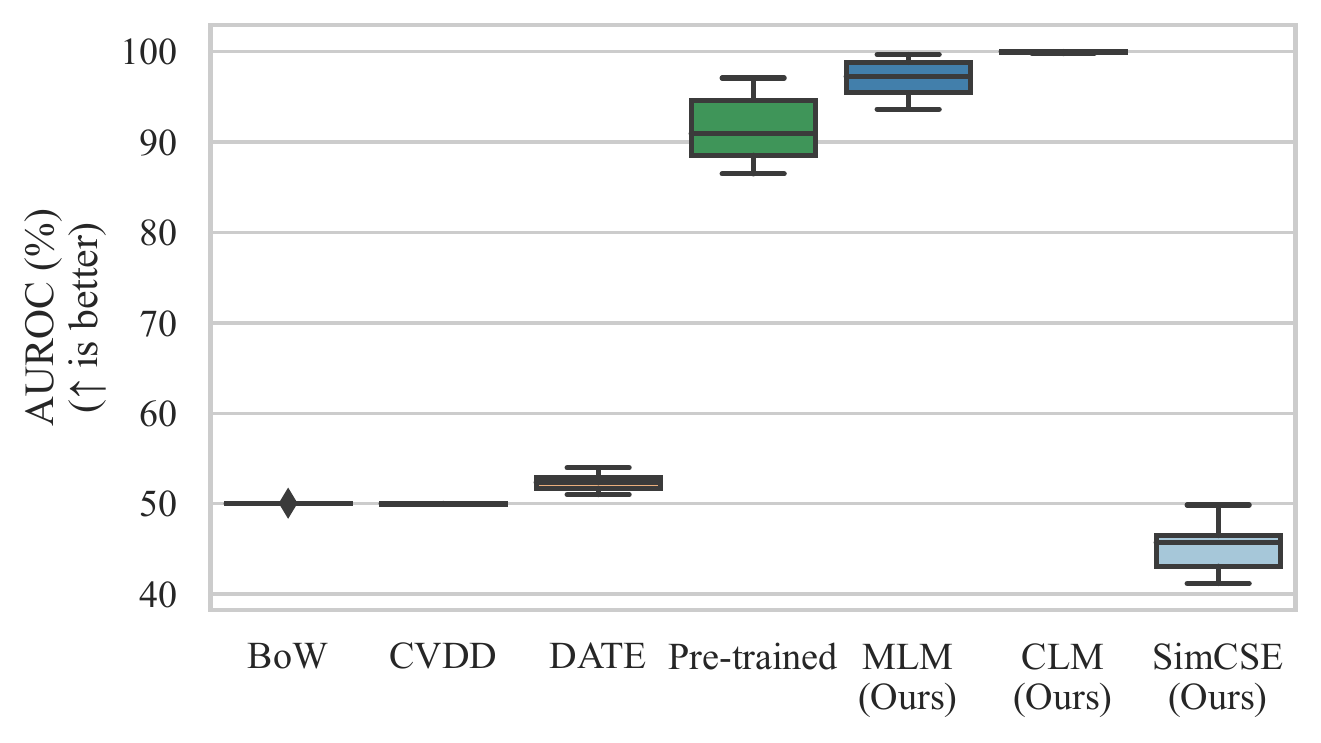} }}\\
    \caption{Syntactic anomaly detection results split by dataset. The figures include all $n$-gram runs.}%
    \label{fig:syntactic results}%
\end{figure*}

\twocolumn
\subsection{Contamination results}
%tbc - need to decide whether to include this or not
%tbc - complete
\begin{figure}[h]
    \centering
    \includegraphics[width=\linewidth]{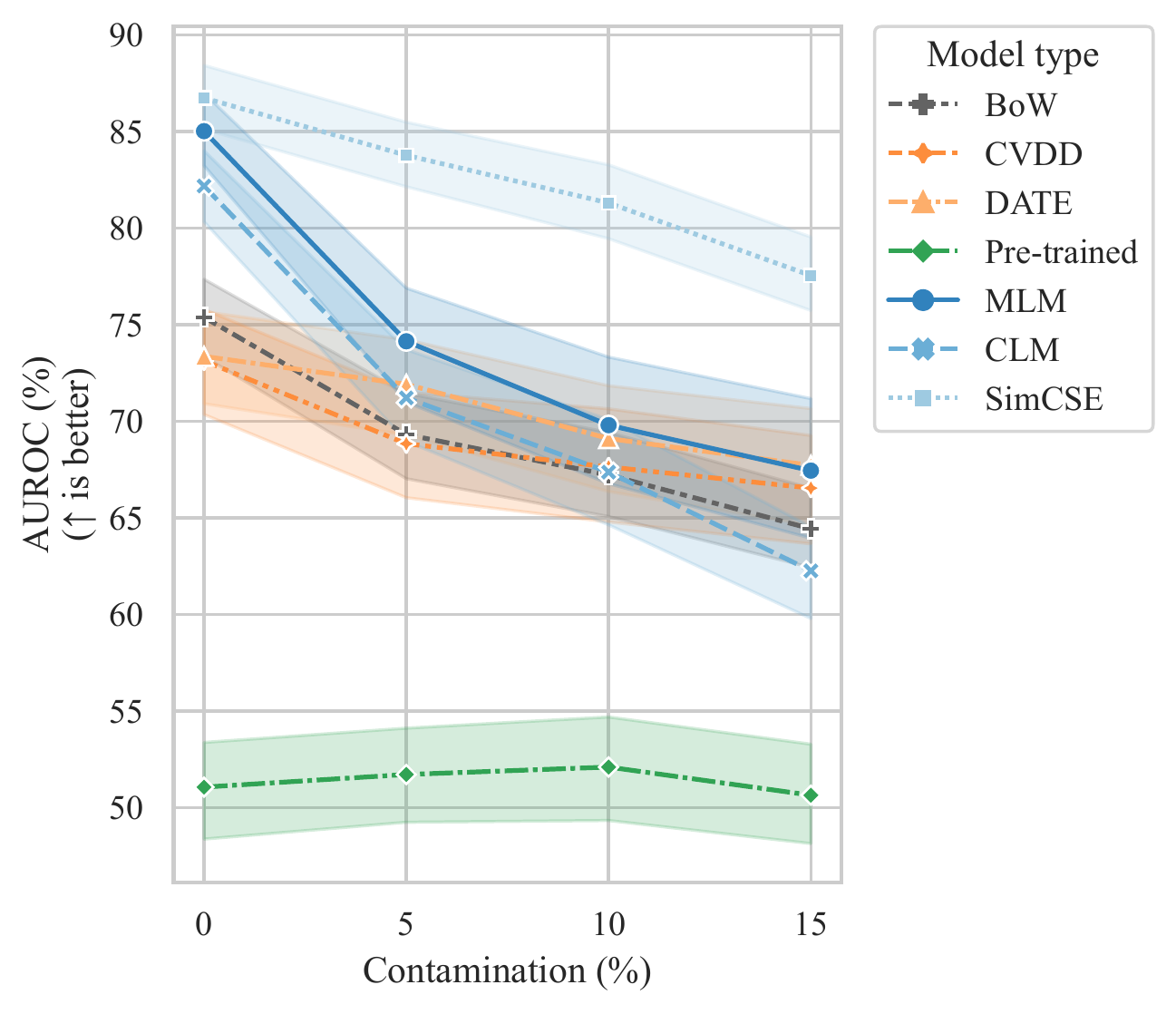}
    \caption{Mean AUROC scores across datasets by contamination percentage. Experiments are conducted using semantic anomalies.}
    \label{fig:contamination}
\end{figure}
We simulate a purely unsupervised anomaly detection setup by incorporating a set percentage of semantic anomalies $\{5\%, 10\%, 15\%\}$ into the training data. The self-supervised losses on average elicit higher AUROC scores compared to the other model types, and SimCSE appears to be the most robust approach.

\subsection{Implementation details}
We used an NVIDIA RTX Titan X and NVIDIA Tesla V100s to run our experiments depending on availability.\\

\noindent\textbf{Model implementation.} We used \texttt{Huggingface}'s\footnote{\url{https://huggingface.co}} implementation of BERT\textsubscript{BASE} and \texttt{Sentence-Transformers}\footnote{\url{https://sbert.net}} for our Transformer experiments. In addition, we used \texttt{nltk}\footnote{\url{https://nltk.org}} for pre-processing, \texttt{spaCy}\footnote{\url{https://spacy.io}} for encoding the bag-of-words models, \texttt{Faiss}\footnote{\url{https://faiss.ai}} to train the $k$-NNs, and \texttt{sci-kit learn}\footnote{\url{https://scikit-learn.org}} for constructing OC-SVMs. \\

\noindent\textbf{Dataset details.} All of the datasets used in our paper are publicly available.

\begin{itemize}
    \item 20 Newsgroups \citep{lang95} is a collection of 20,000 newsgroup documents split across 20 different newsgroups. We use the six top-level subjects (\textit{computer, recreation, science, miscellaneous, politics, religion}) to partition the classes. Partitioning by class label, there are 577-2859 training samples and 382-1909 test samples.
    \item Reuters-21578 \cite{Lewis1997Reuters21578TC} is a collection of 10,788 news articles split across 90 topics. We only use a subset of data that have only one label (\textit{earn, acq, crude, trade, money-fx, interest, ship}). Partitioning by class label, there are 108-2,840 training samples and 36-1,083 testing samples.
    \item AG News \cite{Zhang2015CharacterlevelCN} is a topic classification dataset gathered from more than 2,000 news sources over one year of activity. It contains four classes (\textit{business, sci, sports, world}), each with 30,000 samples for training and 1,900 for testing.
    \item IMDb \cite{maas-EtAl:2011:ACL-HLT2011} is a sentiment classification dataset consisting of film reviews. It contains two classes (\textit{pos, neg}), each with 25,000 samples for training and 25,000 for testing.
    \item Snopes \cite{vo2020facts} is a fact-checking dataset containing paired examples of tweets and a fact-checking article from \textit{snopes.com}. There are four classes (\textit{true, mostly true, mostly false, false}). We only use \textit{true} (7,363) and \textit{false} (21,256) tweets in our experiments and do not use the articles. We randomly partition 80\% of this smaller dataset for training and use the remaining 20\% for testing.
    \item The Enron Spam Dataset \cite{metsis2006} is derived from the Enron Email Dataset. There are two classes, \textit{ham} (16,458) and \textit{spam} (17,171) emails. We randomly partition 80\% of the dataset for training and the remaining 20\% for testing.
    
\end{itemize}

\end{document}